\documentclass[a4paper]{llncs}	
\usepackage{graphicx}
\usepackage[latin1]{inputenc}
\usepackage{float}

\usepackage{subcaption}
\begin{document}
\title{DFKI Cabin Simulator: A Test Platform for Visual In-Cabin Monitoring Functions}

\author{Hartmut Feld, Bruno Mirbach, Jigyasa Katrolia, Mohamed Selim, Oliver Wasenm\"{u}ller\and Didier Stricker}
\institute{German Research Center for Artificial Intelligence (DFKI),\\ Trippstadter Stra{\ss}e 122, 67663 Kaiserslautern
	\email{\\
		\{firstname.lastname\}@dfki.de  \\http://av.dfki.de} }

\maketitle

\begin{abstract}
We present a test platform for visual in-cabin scene analysis and occupant monitoring functions.
The test platform is based on a driving simulator developed at the DFKI, consisting of a realistic 
in-cabin mock-up and a wide-angle projection system for a realistic driving experience.
The platform has been equipped with a wide-angle 2D/3D camera system monitoring the entire interior of the vehicle mock-up of the simulator. 
It is also supplemented with a ground truth reference sensor system that allows to track and record 
the occupant's body movements synchronously with the 2D and 3D video streams of the camera.
Thus, the resulting test platform will serve as a basis to validate numerous in-cabin monitoring functions, 
which are important for the realization of novel human-vehicle interfaces, advanced driver assistant systems, 
and automated driving. Among the considered functions are occupant presence detection, size and 3D-pose estimation, and driver intention recognition. 
In addition, our platform will be the basis for the creation of large-scale in-cabin benchmark datasets.
\end{abstract}

\begin{keywords}
In-Cabin Monitoring, Driving Simulator, Pose Estimation, Benchmark Dataset Creation
\end{keywords}

\section{Introduction}
In-cabin monitoring of vehicle occupants is a topic of increasing interest induced by the ongoing development
of advanced driver assistant systems (ADAS) and automated driving, up to driverless vehicles and shared mobility. The requirements towards the monitoring functions are thereby manifold and changing with the level of driving automation. The demand for these novel monitoring functions comes moreover from new requirements for safety and comfort functions, as well as from novel human-vehicle interfaces.

The full monitoring and understanding of the scene in the vehicle cabin comprises not only the automatic detection, classification, and recognition of all occupants and objects, but also the estimation of the occupants' pose and state, as well as the recognition of their activities, interactions, and intentions.

Of particular interest is thereby the monitoring of the driver's state and intention. Many research activities have recently concentrated on the development of camera systems monitoring the driver's face and inferring on his state such as awareness and focus of attention, to realize novel warning functions and as support for ADAS functions and future automated driving.

Although full-body pose tracking of humans has become an intensive research domain since the launch of the first Kinect camera \cite{Habibie_2019_CVPR,shotton2011real-time}, the full-body pose detection in vehicles is rarely investigated. The benefits of monitoring the full-body pose in a vehicle are thereby manifold. 
They range from the recognition of arm gestures and intentions for comfort and advanced human-vehicle interfaces
to a robust analysis of the driver's activities and availability. The latter function is crucial for automated driving of levels 3 and 4, in which the hand-over of the vehicle control to the driver has to be managed.

Existing benchmark datasets for in-cabin monitoring functions, as, e.g. the VIVA challenge \cite{VIVA},
often provide videos of confined areas inside the vehicle, where either the driver's head or hand is located.
Moreover, the annotations do not contain 3D ground truth data. Recently, several in-cabin benchmark datasets with precise ground truth 3D-measurements have been published \cite{Roth_IV2019,Schwarz_2017_CVPR_Workshops}, but they are restricted to the detection of the head pose. There exist full-body pose datasets, as, e.g. the MPII dataset \cite{andriluka14cvpr}. This benchmark contains a very wide range of scenes of which only a few are "driving automobile".

In this work, we go further by developing a test platform that allows recording large-scale datasets of complete in-cabin scenes. In this way, we close the gap between systems monitoring only the head or hands of the driver. Recorded data will comprise annotations for a wide range of monitoring functions, including also ground truth 3D body pose measurements of the vehicle occupants.

\section{The DFKI in-cabin test platform}
\label{s:testplatform}

The in-cabin test platform is based on a driving simulator developed at the DFKI, consisting of a 
realistic in-cabin mock-up and a wide-angle projection system for a realistic driving experience. 
The test platform has been equipped with a wide-angle 2D/3D camera system for monitoring the entire
interior of the vehicle mock-up and an optical ground truth reference sensor system that allows to track and record the occupant's body movements synchronously with the 2D and 3D video streams of the camera. Moreover, the precise positioning of the front seats can be controlled and registered via a CAN-interface.

\begin{figure}[H]
	\begin{center}
		\includegraphics[width=0.8\textwidth]{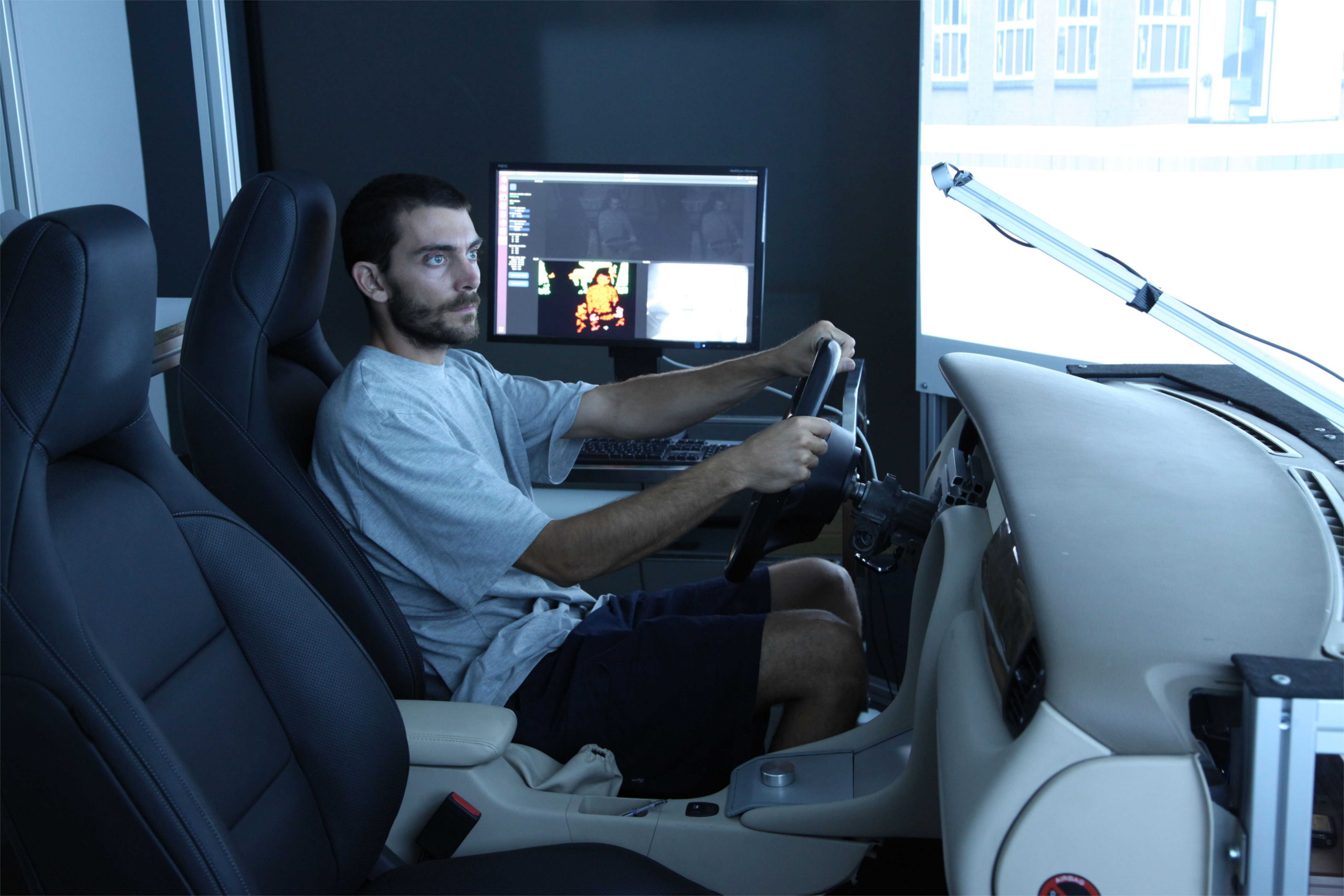}
		\caption{Driving actions recorded with a 2D/3D camera system mounted above the dashboard. 
			The simulated street scene can be recognized on the right;
			the computer screen in the background displays the data stream from the camera.}
		\label{fig:driving}
	\end{center}
\end{figure}

Fig. \ref{fig:driving} shows the testing process in action. In addition to the three-screen projection of the simulator software, another monitor is mounted on the side of the simulator for the test engineer in order to control the recording or testing process. With this setup, it is possible to visualize data while recording and to test and demonstrate functionalities in real-time. The individual components of the test setup are described in more detail in the following sub-sections.

\begin{figure}
	\centering
	\begin{subfigure}[b]{0.475\textwidth}
		\centering
		\includegraphics[width=\textwidth]{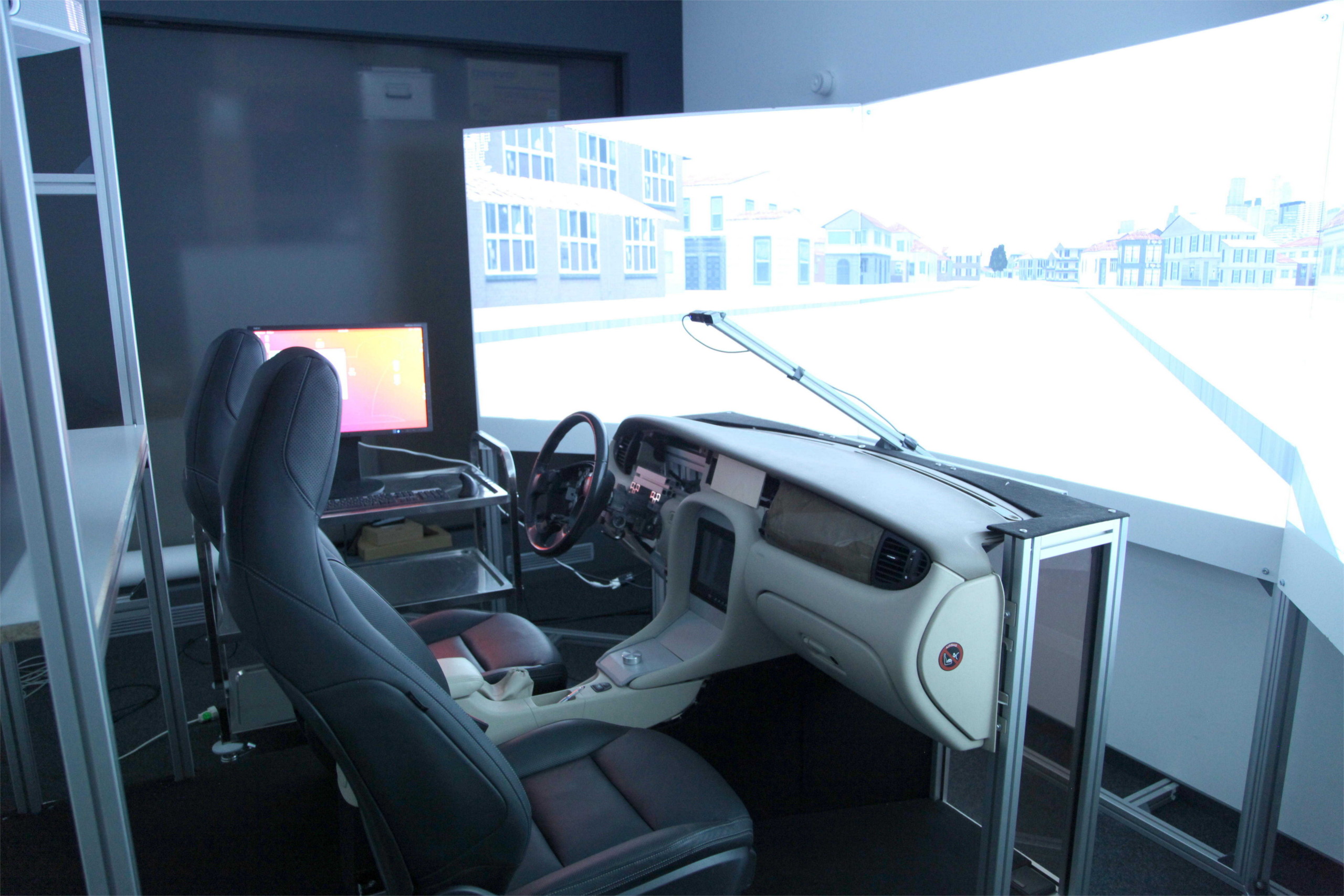}
		\caption{Full view with running simulation}
		\label{fig:full}
	\end{subfigure}
	\begin{subfigure}[b]{0.475\textwidth}
		\centering
		\includegraphics[width=\textwidth]{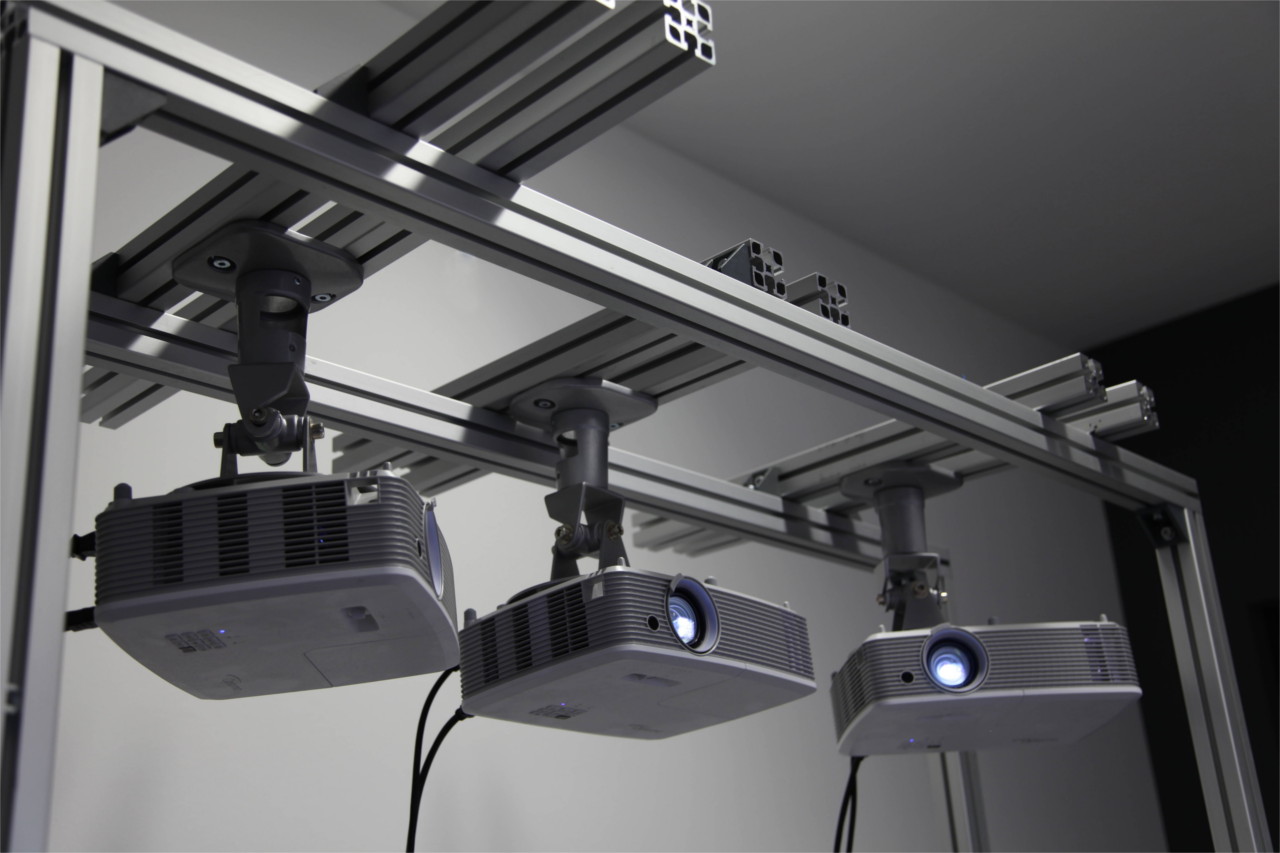}
		\caption{Simulation projectors}
		\label{fig:projectors}
	\end{subfigure}
	\begin{subfigure}[b]{0.475\textwidth}
		\centering
		\includegraphics[width=\textwidth]{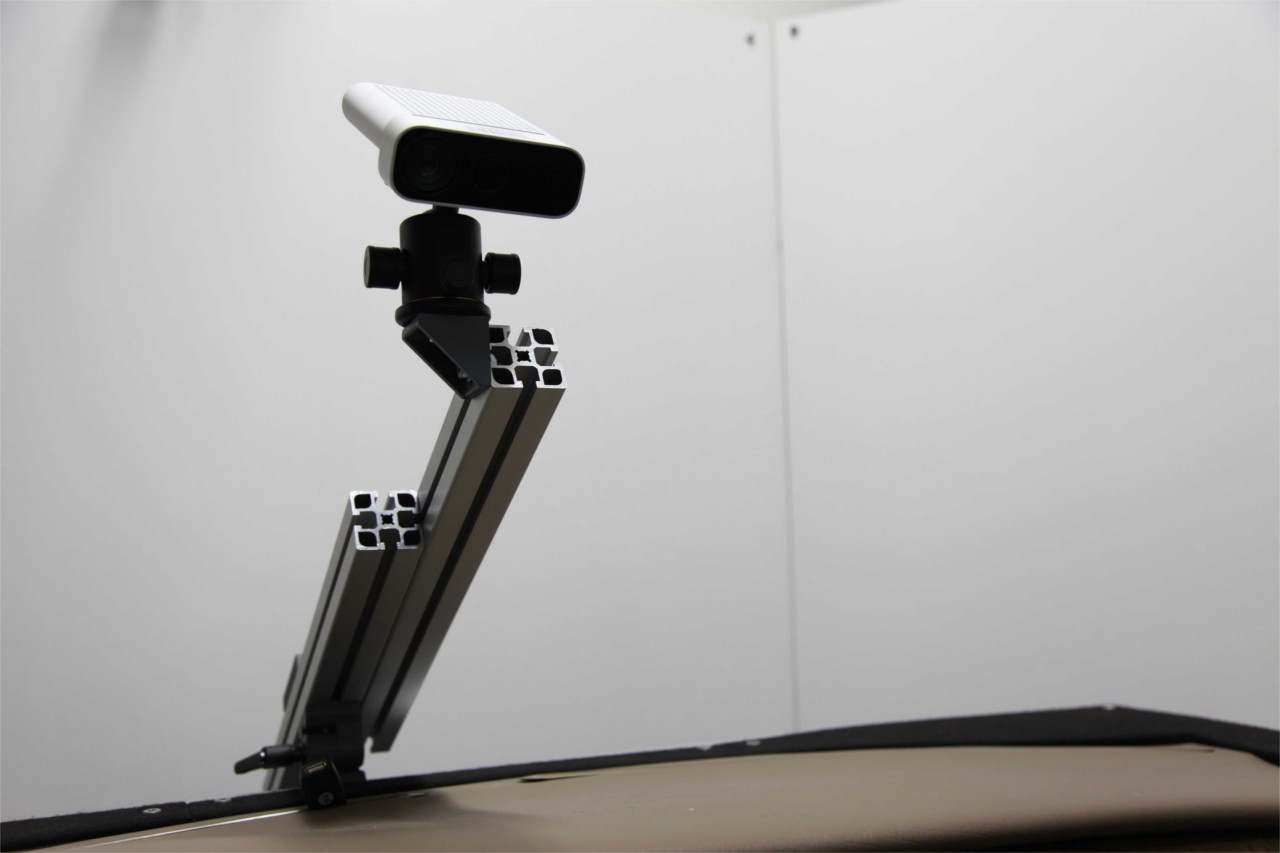}
		\caption{Camera mounting front view}
		\label{fig:cam_mounting1}
	\end{subfigure}
	\begin{subfigure}[b]{0.475\textwidth}
		\centering
		\includegraphics[width=\textwidth]{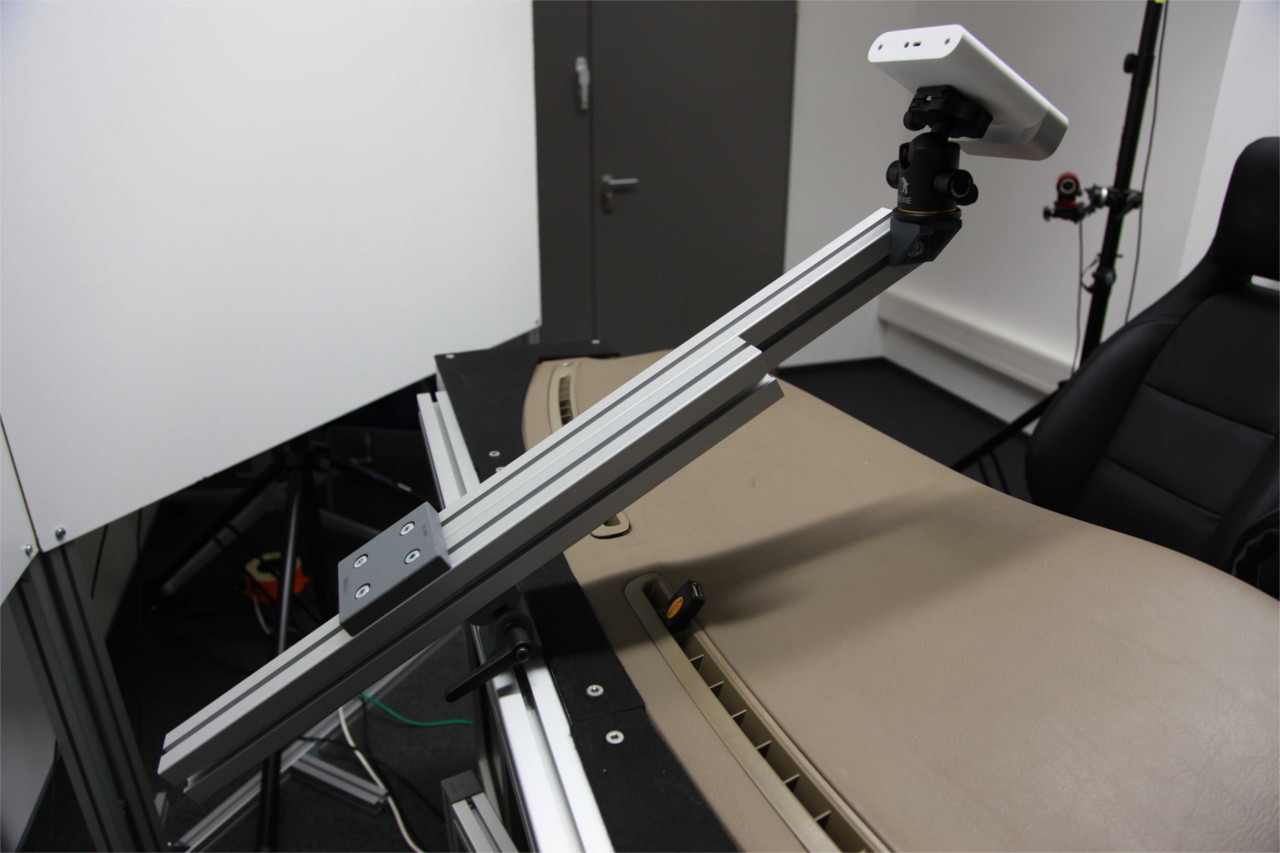}
		\caption{Camera mounting back view}
		\label{fig:cam_mounting2}
	\end{subfigure}
	\begin{subfigure}[b]{0.475\textwidth}
		\centering
		\includegraphics[width=\textwidth]{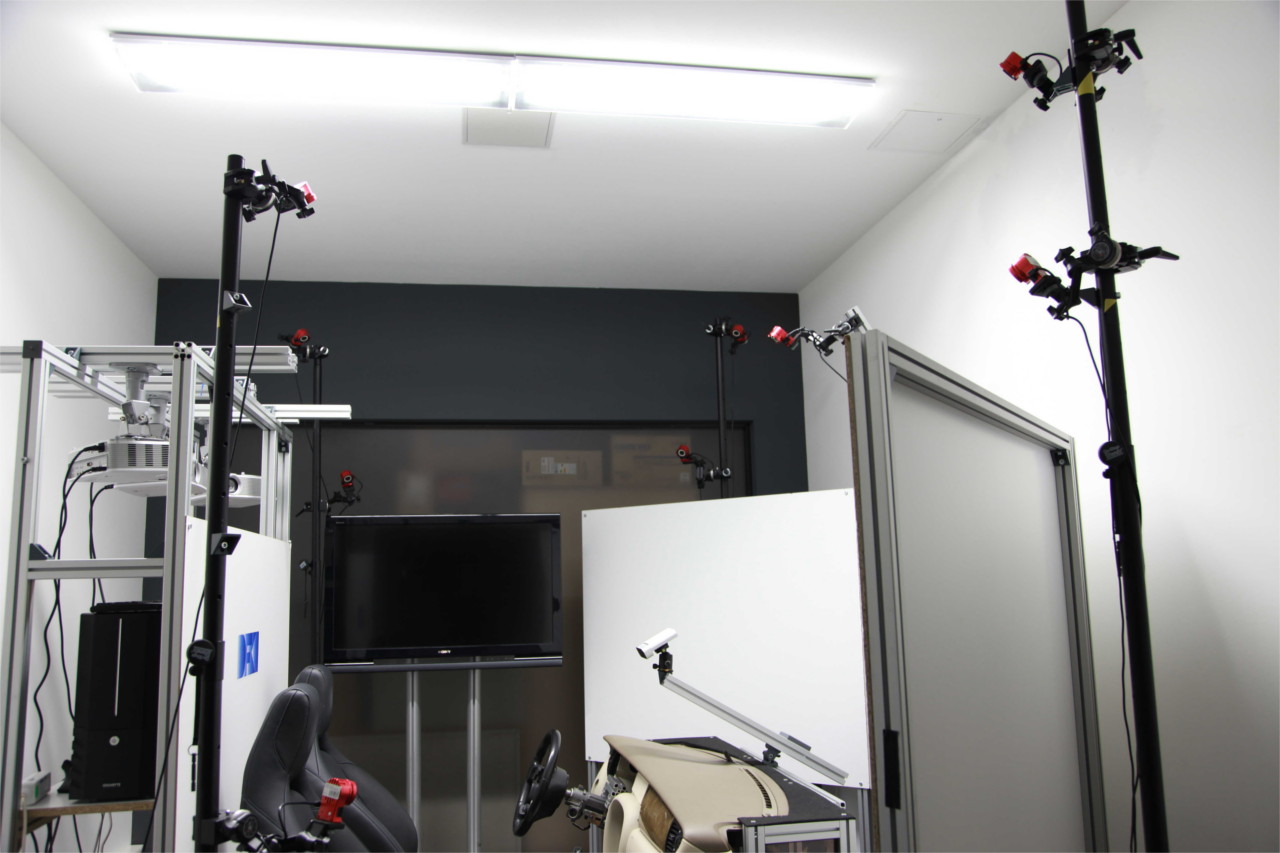}
		\caption{Ground truth sensor system}
		\label{fig:optitrack1}
	\end{subfigure}
	\begin{subfigure}[b]{0.475\textwidth}
		\centering
		\includegraphics[width=\textwidth]{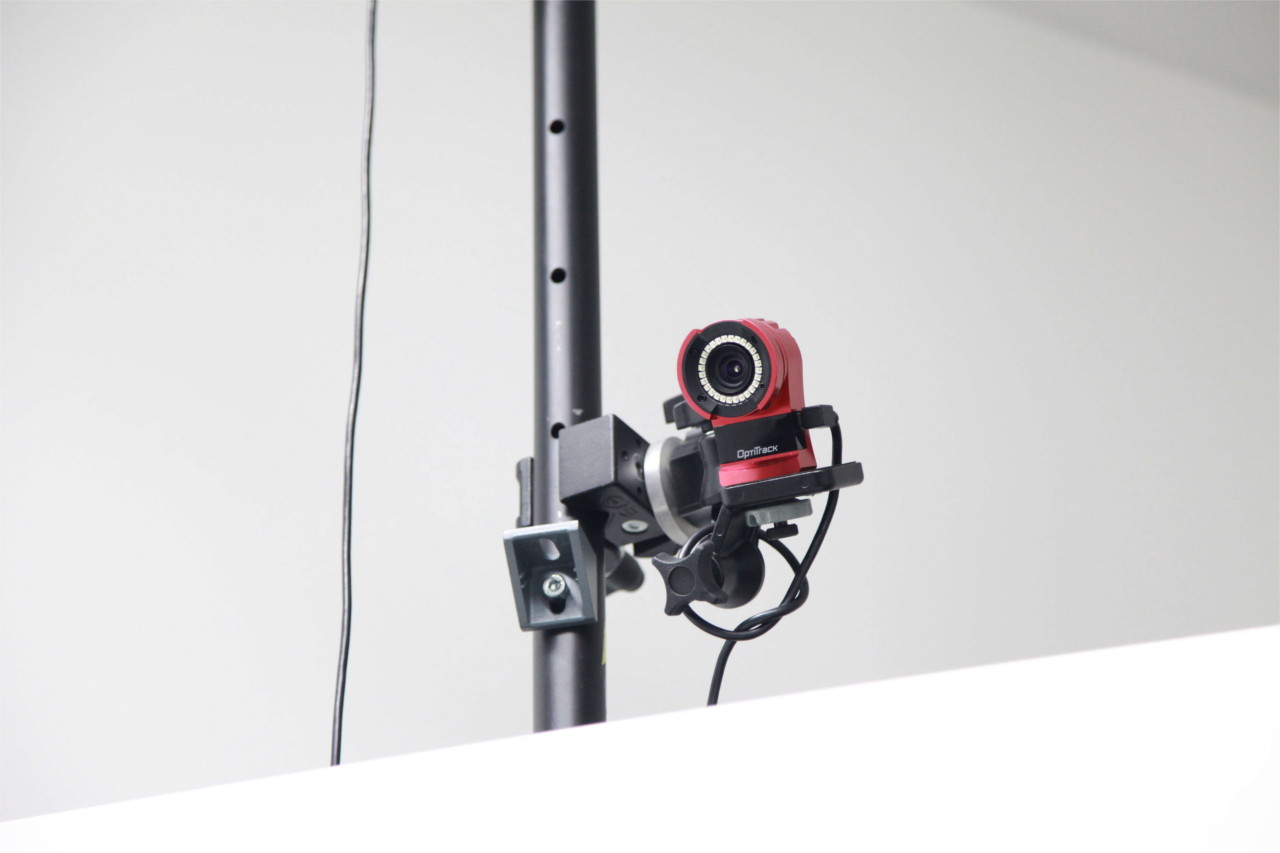}
		\caption{Ground truth sensor camera}
		\label{fig:optitrack2}
	\end{subfigure}
	\begin{subfigure}[b]{0.475\textwidth}
		\centering
		\includegraphics[width=\textwidth]{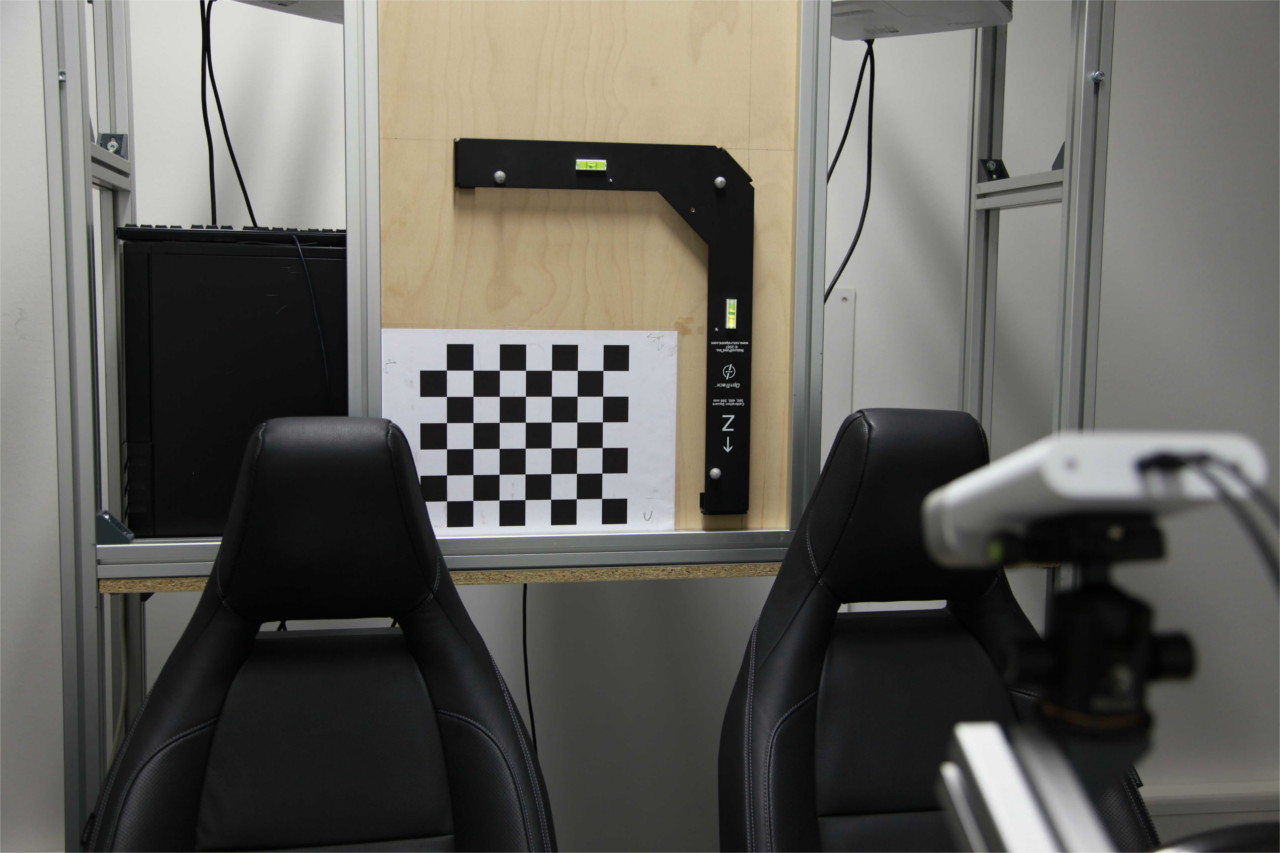}
		\caption{Mounting of calibration targets}
		\label{fig:calib}
	\end{subfigure}
	\begin{subfigure}[b]{0.475\textwidth}
		\centering
		\includegraphics[width=\textwidth]{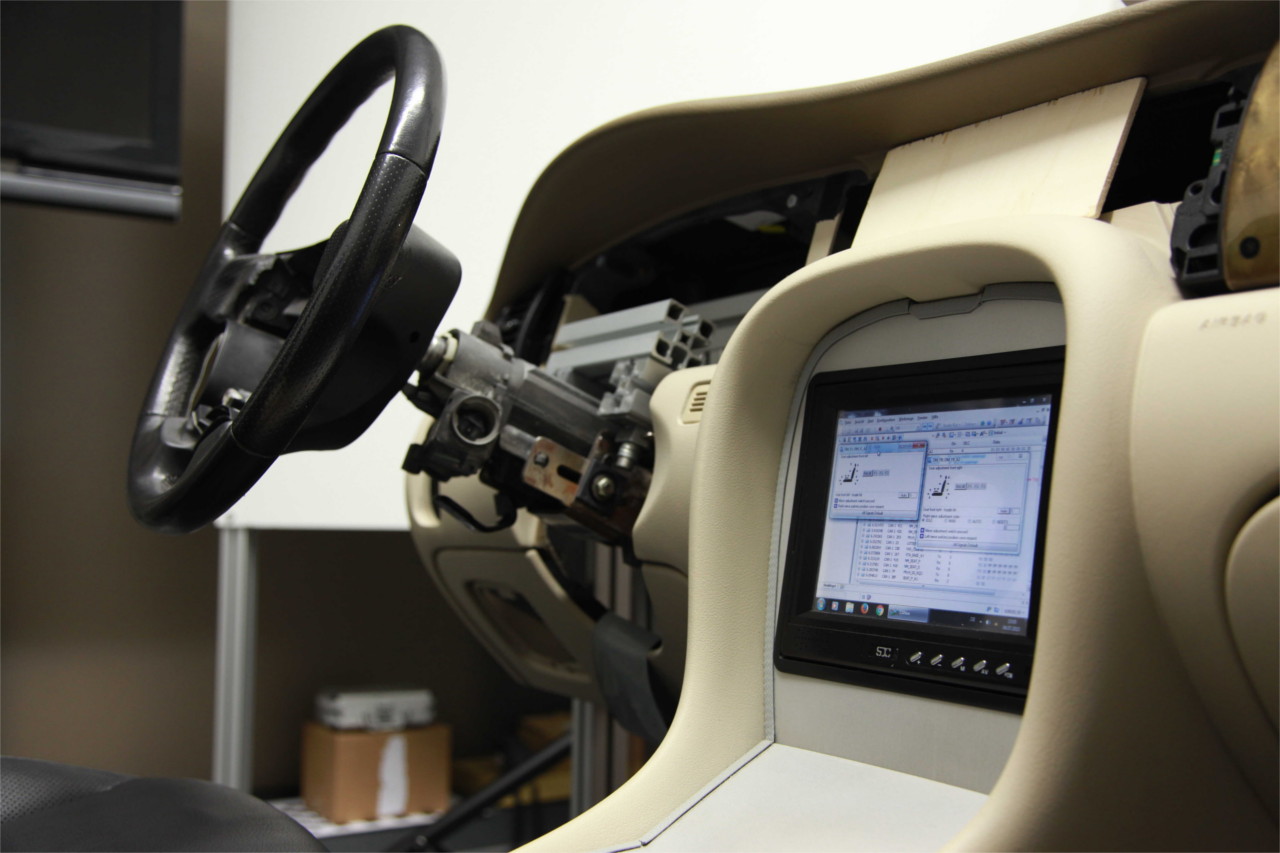}
		\caption{Seat positioning software}
		\label{fig:seat_monitor}
	\end{subfigure}
	
	\caption{Images of the driving simulator.}
	\label{fig:ds_all}
\end{figure}

\subsection{Driving Simulation}
The OpenDS simulator software \cite{opends} is utilized and provides a near-realistic driving environment. There are over 20 different driving scenarios and driving tasks in urban and rural environments including other traffic participants like cars and pedestrians as well as traffic lights. The software is also capable of simulating different environmental and weather conditions.

Three projectors (Fig. \ref{fig:projectors}) are used to project the simulated scene on a wide-angle screen covering almost the driver's entire field of vision (Fig. \ref{fig:full}). This leads to a realistic driving experience and, most importantly, to realistic driver's movements (e.g. head movement in turning maneuvers). Together with the simulation, the software also comprises a drive analyzer to gather the data of the USB Controller that senses the driving actions as steering, braking, and acceleration.

Of course, it is also possible to install other simulation software. This might come in handy to cover special needs for testing e.g. commercial vehicles, agricultural vehicles.

\subsection{2D/3D camera system mounting}
\label{ss:cam_mounting}
The test platform has been equipped with a wide-angle 2D/3D camera system for monitoring the entire interior of the vehicle mock-up of the simulator. The mounting position of the camera corresponds to the overhead module of the car close to the rear mirror (see Fig. \ref{fig:cam_mounting1} and \ref{fig:cam_mounting2}). In order to be capable of monitoring both driver and passenger from that position, a wide field of view to monitor the full vehicle cabin, including the driver and the front-seat passenger is crucial. The metal fixation bar also allows mounting several cameras in parallel in similar positions. Section \ref{s:camera} discusses several 2D/3D camera systems that had been mounted and evaluated in this test setup. 

Due to the multi-purpose aluminum profile framework, the mounting is, however, also flexible enough to mount 
additional cameras and other sensors anywhere according to various specific needs.

\subsection{Ground truth sensor system}
\label{ss:optitrack}
\begin{figure}[h!]
	\begin{center}
		\includegraphics[width=0.8\textwidth]{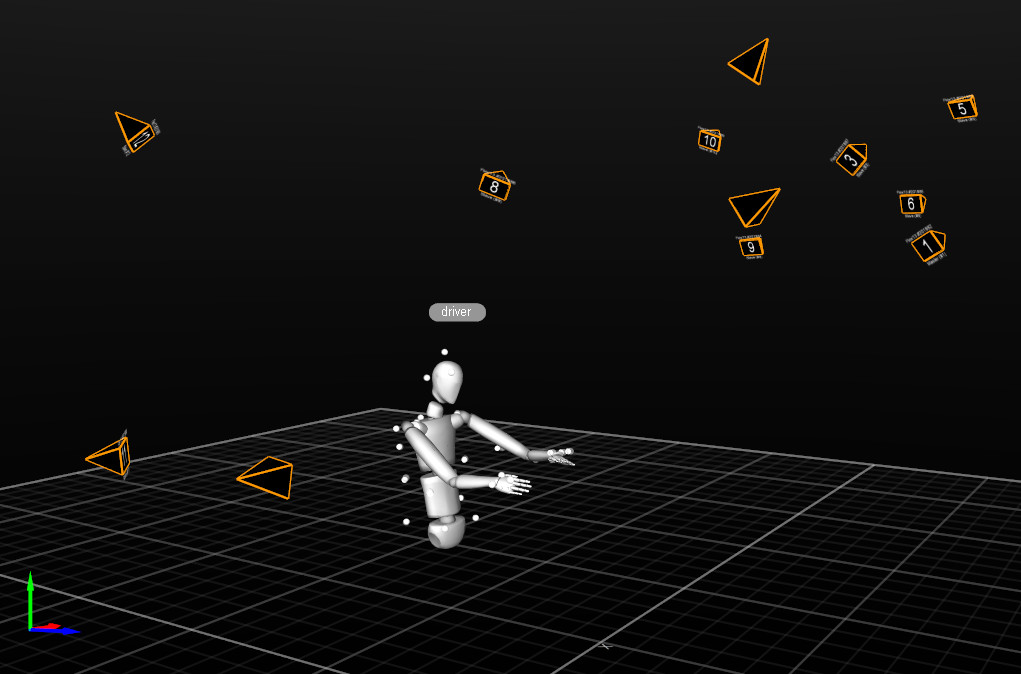}
		\caption{Driving actions recorded with OptiTrack \cite{optitrack} sensor system to extract ground truth joint positions. 
			The rectangular pyramids indicate the position and rotation of the 12 cameras in the room. The detected 3D positions
			of the markers is shown as white dots around a schematic skeleton.}
		\label{fig:opti_skeleton}
	\end{center}
\end{figure}

This test platform is supplemented with a ground truth sensor system that allows to track and record the occupant's movement at high frame rates, synchronously with the 2D and 3D video stream of the camera system monitoring the cabin of the driving simulator. The 3D tracking system chosen for that purpose is OptiTrack \cite{optitrack}. OptiTrack is a motion capture system that works with IR cameras detecting small reflective markers on the subject's body joints. It has often been used as a reference system in the past (e.g.  \cite{wasenmueller2016corbs}).
Fig. \ref{fig:optitrack1} and \ref{fig:optitrack2} show the OptiTrack cameras mounted around the driving simulator in a way that each tracking point on the subject is visible at all times. 
With this tracking mechanism, it is possible to record the position of each joint for every point in time with sub-millimeter accuracy. This leads to an easy and precise automated testing of algorithms. Fig. \ref{fig:opti_skeleton} shows a skeleton preview of the OptiTrack software.

\subsection{Sensor Calibration and Synchronization}
A metal frame holding the simulator projectors is rigidly fixed to the vehicle mock-up and defines the so-called world coordinate system. Fig. \ref{fig:calib} shows the designated calibration board that can be mounted on the simulator frame. It offers a checkerboard for camera calibration as well as a calibration square for the ground truth sensor system. This setup enables extrinsic calibration of both the installed camera and the ground truth sensor system in relation to the world coordinate system. The precisely mounted board solution guarantees precise calibration and position verification. The calibration board can be unmounted from the frame to make way for undisturbed data recording.

Synchronization between the camera and ground truth sensor system is realized via a synchronization wire. The ground truth system sends periodic pulses with exposure frequency as soon as data recording is started. The camera is triggered with these pulses and gets re-synchronized to avoid drift.

\subsection{Seat positioning}
\label{ss:seatpos}
\begin{figure}
	\begin{center}
		\includegraphics[width=0.8\textwidth]{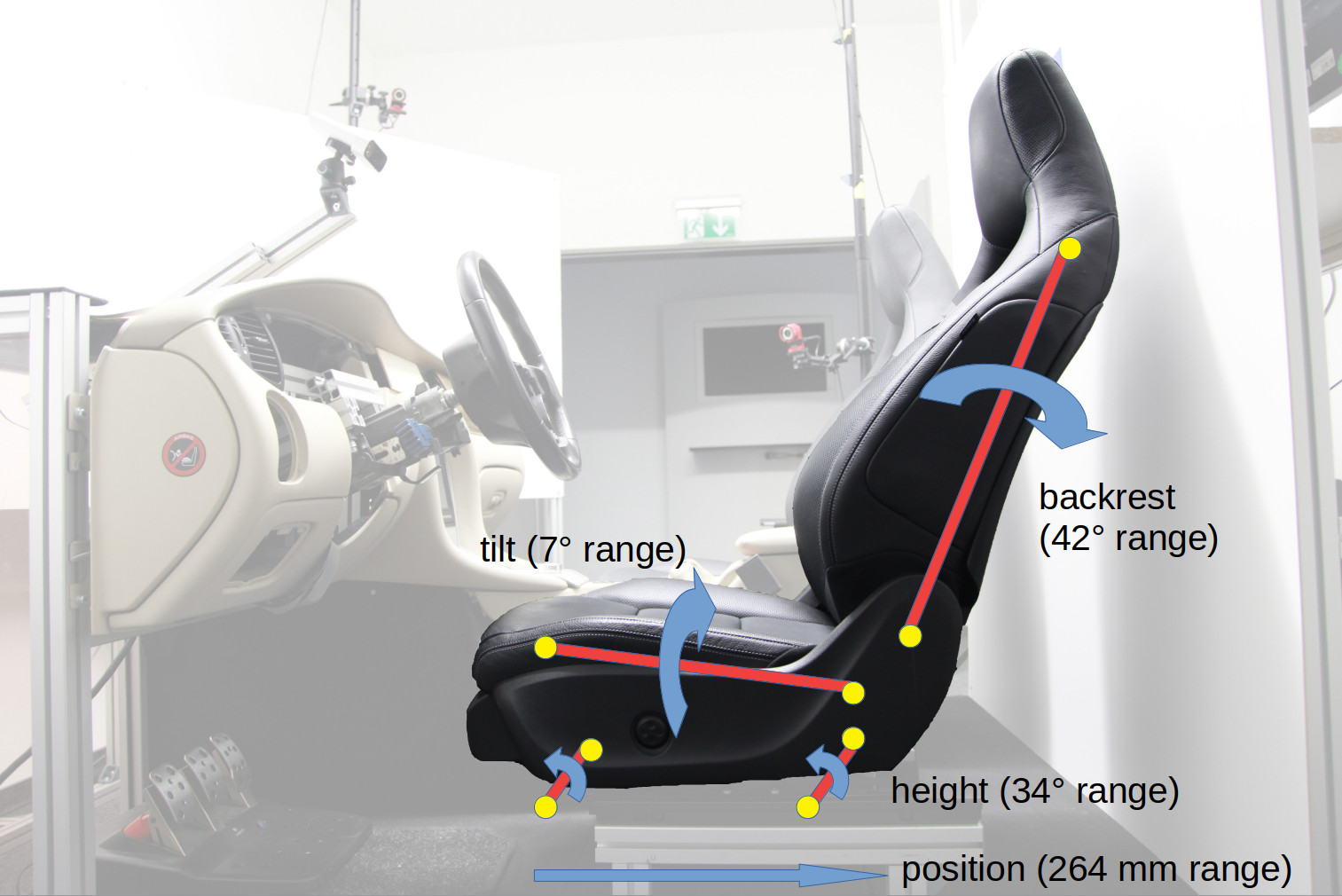}
		\caption{Variables for the seat positioning mechanism.}
		\label{fig:seat_pos_mech}
	\end{center}
\end{figure}

The positioning of the driver and passenger seat in the mock-up is controlled via the CAN bus system of the seats which originate from a real automobile (see Fig. \ref{fig:seat_monitor}). This access via CAN guarantees the full traceability of the seat position while testing and also enables a precise seat adjustment according to predefined positions. Another very important aspect of the testing or development of deep learning algorithms is to test and train them with a wide variability of data. For depth imagery, the most interesting and realistic way to add data variability to the depth data is by changing the seat position. The CAN control of the seats provides the full range of seat positions with a set of four degrees of freedom: seat height, seat position, backrest position, and seat tilt. Fig. \ref{fig:seat_pos_mech} schematically shows these variables.
In addition to that, the steering wheel position is also alterable, though not controlled via the CAN-interface.
With that variability, it is possible to cover the full range of possible real-world seat positions by a comprehensive pre-defined set of parameter combinations. The influence of the seat pose variation on the in-cabin scenery recorded by the selected 2D/3D-camera is discussed further below in Section \ref{s:camera} (see also Fig. \ref{fig:seat_extr}).

\section{In-cabin monitoring with a 2D/3D camera system}
\label{s:camera}
\subsection{Selection of a 2D/3D camera system for in-cabin monitoring}
In order to monitor the entire interior of the vehicle mock-up from a mounting position corresponding to the overhead module of the car close to the rear mirror (see Section \ref{ss:cam_mounting}), a camera field of view of at least 120° is required. The depth-sensing range has to cover a range of 25cm to 200cm.
Also, the camera needs to capture data at a high framerate of at least 30Hz to be capable to track person's movements. 

Several depth cameras available on the market have been evaluated against these requirements. 
Tab. \ref{table:cam_tab} summarizes the main optical characteristics of the reviewed cameras.

\begin{table}[h!]
\begin{center}
	\begin{tabular}{ |p{3cm}||p{3cm}|p{3cm}|p{3cm}|  }
		\hline
		Camera Name & Type & Depth Field of View & Depth Range in m \\
		\hline
		\hline
		Kinect v1 \cite{kinectv1}  & IR Pattern and RGB    & 57 x 43 &   1.2 - 3.5 \\
		Kinect v2 \cite{kinectv2} &   ToF and RGB  & 70 x 60   & 0.5 - 4.5 \\
		Azure Kinect \cite{azure} & ToF and RGB & 120 x 120 &  0.25 - 2.88 \\
		MYNT EYE S \cite{mynt} & Stereo + IR Pattern & 120 x 75 & 0.7 - 3 (IR range)\\
		Stereolabs ZED \cite{zed} & Stereo RGB & 90 x 60 & 0.5 - 20\\
		Structure Core \cite{structure} & IR Pattern and RGB & 58 x 45 & 0.4 - 5\\
		\hline
	\end{tabular}
	\caption{Reviewed cameras}
	\label{table:cam_tab}
\end{center}
\end{table}

\begin{figure}
	\centering
	\begin{subfigure}[b]{0.35\textwidth}
		\centering
		\includegraphics[width=\textwidth]{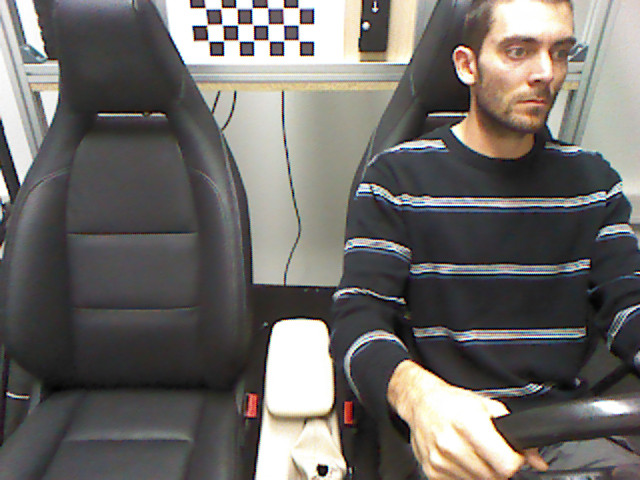}
		\label{fig:kinect_v1_rgb}
	\end{subfigure}
	\begin{subfigure}[b]{0.35\textwidth}
		\centering
		\includegraphics[width=\textwidth]{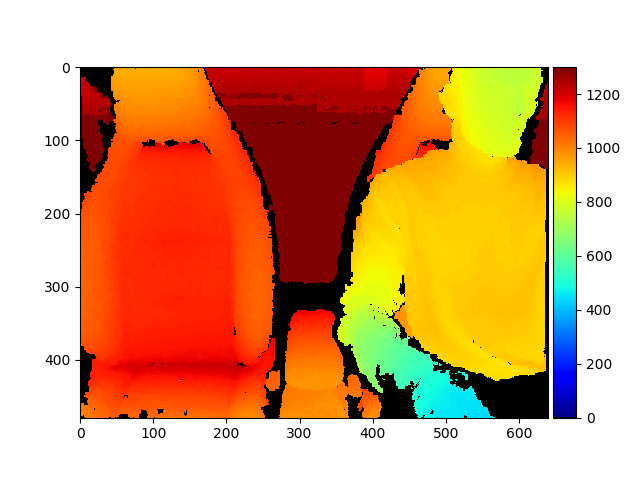}
		\label{fig:kinect_v1}
	\end{subfigure}
	\begin{subfigure}[b]{0.35\textwidth}
		\centering
		\raisebox{0.15\height}{\includegraphics[width=\textwidth]{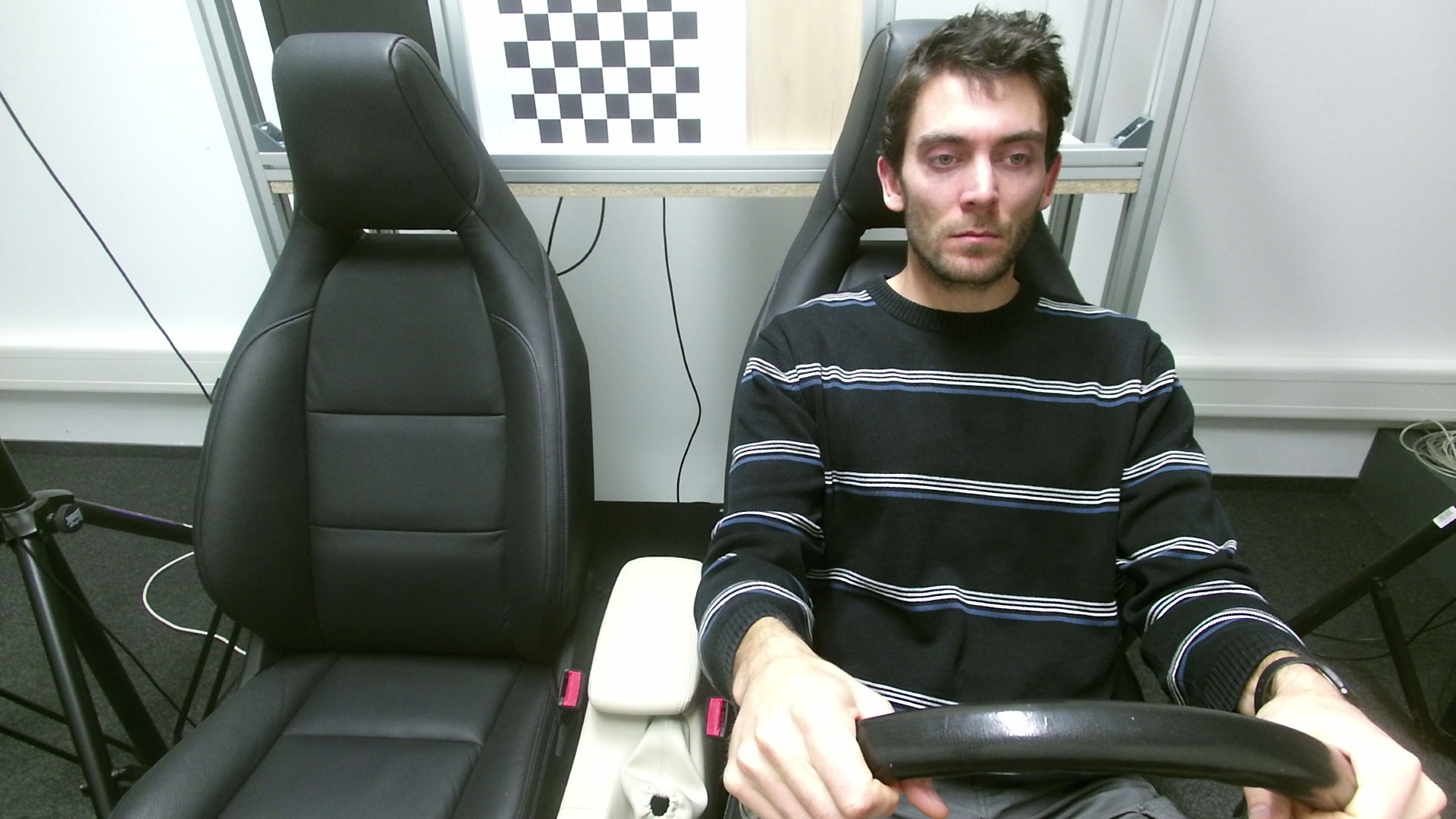}}
		\label{fig:kinect_v2_rgb}
	\end{subfigure}
	\begin{subfigure}[b]{0.35\textwidth}
		\centering
		\includegraphics[width=\textwidth]{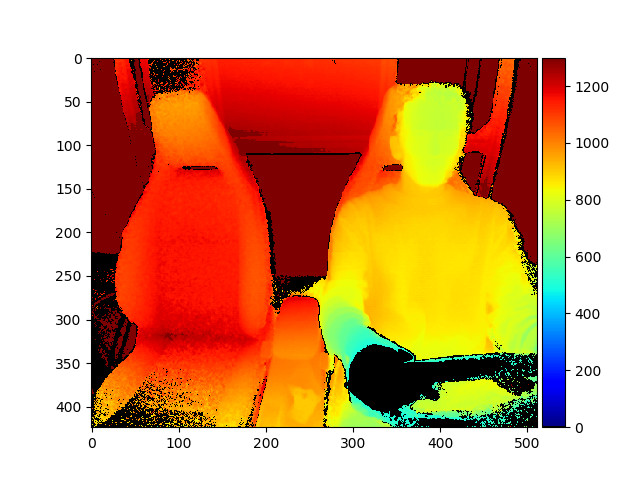}
		\label{fig:kinect_v2}
	\end{subfigure}
	\begin{subfigure}[b]{0.35\textwidth}
		\centering
		\raisebox{0.15\height}{\includegraphics[width=\textwidth]{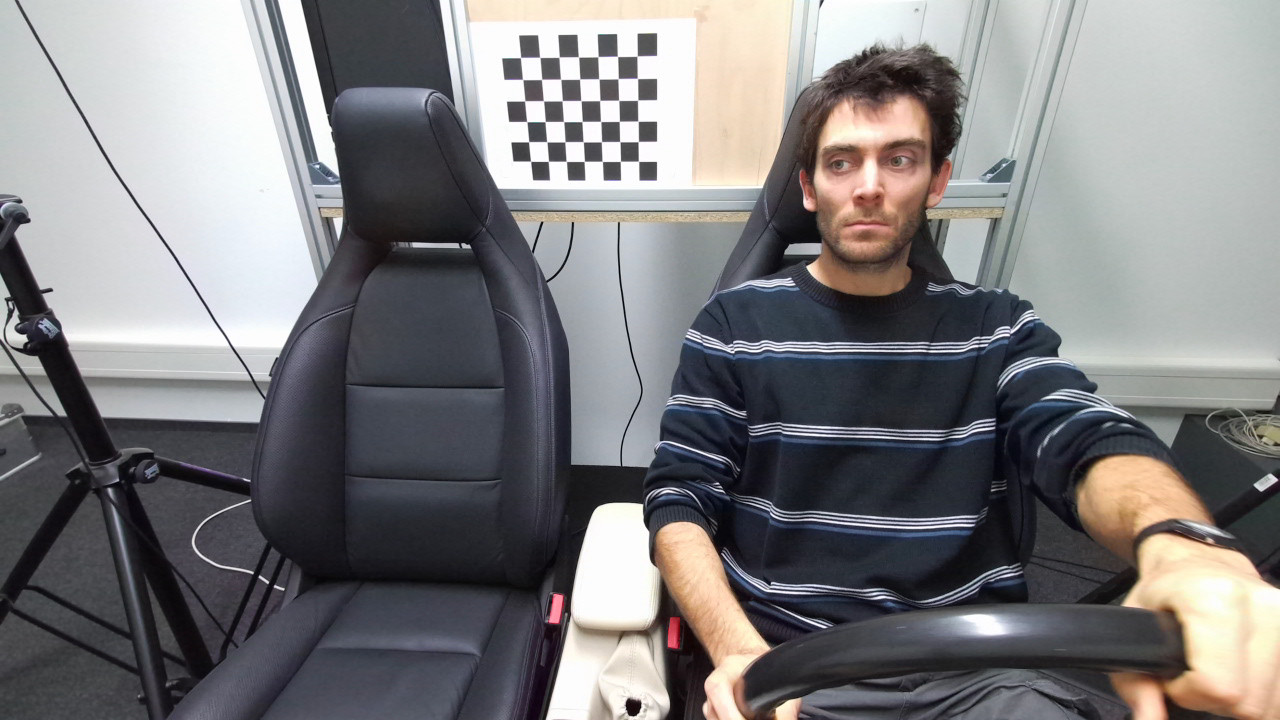}}
		\label{fig:azure_rgb}
	\end{subfigure}
	\begin{subfigure}[b]{0.35\textwidth}
		\centering
		\includegraphics[width=\textwidth]{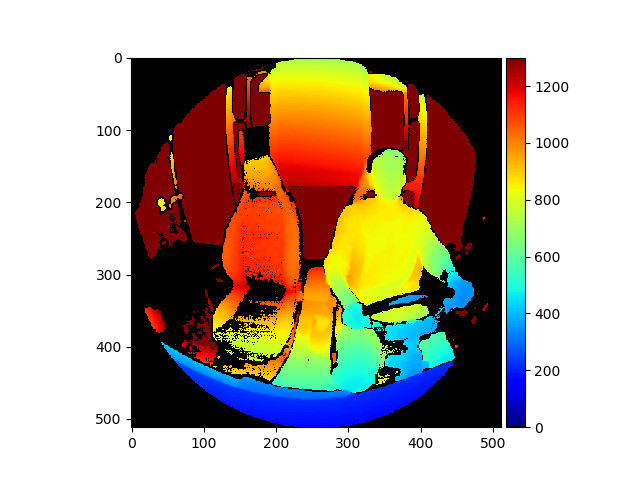}
		\label{fig:azure}
	\end{subfigure}
	\caption{Comparison of depth imagery from three generations of Microsoft Kinect cameras. All images were recorded with the 
	respective cameras in the same position. Depth values are shown in false-color representation where the range was limited to 1300 millimeters for better visualization. From top to bottom: Kinect v1 \cite{kinectv1}, Kinect v2 \cite{kinectv2}, Azure Kinect \cite{azure}.}
	\label{fig:comparison}
\end{figure}

Fig. \ref{fig:comparison} shows example RGB-D images of the various generations of Kinect cameras. The Kinect cameras are consumer cameras also commonly used in the research community, e.g. to record benchmark datasets \cite{firman-cvprw-2016}. The cameras have been mounted on the rear mirror position as described in section \ref{ss:cam_mounting} (see Fig. \ref{fig:cam_mounting1} and \ref{fig:cam_mounting2}).

One recognizes in Fig. \ref{fig:comparison} that only the latest generation Azure Kinect provides a complete view of the cabin interior from the selected rear mirror perspective. The field of view of the Kinect v1 is too small to monitor the driver fully: The left arm is not visible. The Kinect v2 has a larger field of view, 
but the driver's hands are too close to the camera for valid depth measurement. The other available depth sensors show limitations according to the first two generations of the Kinect, as one can already recognize from their datasheets (see Tab. \ref{table:cam_tab}), and were therefore not further considered for the in-cabin test setup.

With these results, we decided that the Azure Kinect is the favorable camera to use. Although it has some limitations, it is the only camera that is capable of recording the whole driver's body. On the downside, the IR transmitter is not strong enough to provide sufficient signal everywhere on the low-remitting black leather seat for valid depth measurement.

\subsection{Influence of seat adjustment on the depth image}
\label{ss:camera_seatpose}
\begin{figure}
	\centering
	\begin{subfigure}{0.33\textwidth}
		\centering
		\includegraphics[height=0.2\textheight]{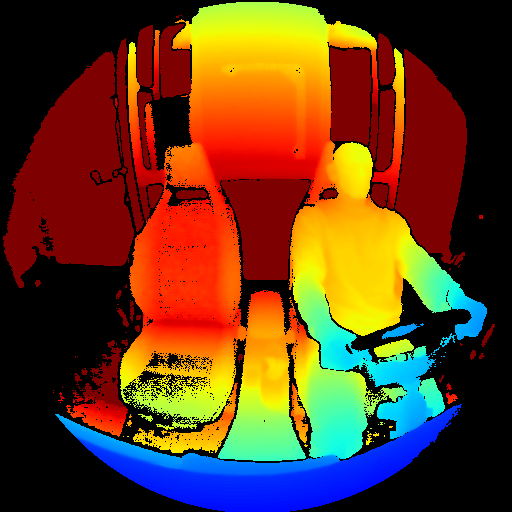}
		\label{fig:b_back}
	\end{subfigure}
	\begin{subfigure}{0.33\textwidth}
		\centering
		\includegraphics[height=0.2\textheight]{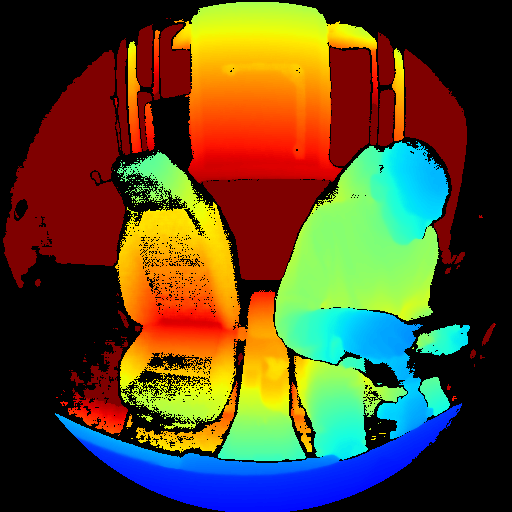}
		\label{fig:b_front}
	\end{subfigure}
	\begin{subfigure}[b]{0.33\textwidth}
		\centering
		\includegraphics[height=0.2\textheight]{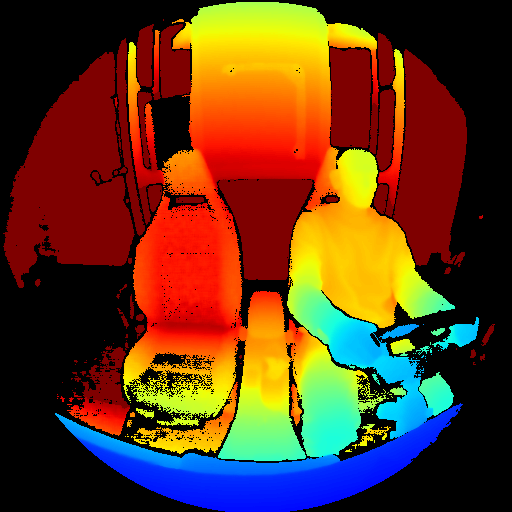}
		\label{fig:h_down}
	\end{subfigure}
	\begin{subfigure}[b]{0.33\textwidth}
		\centering
		\includegraphics[height=0.2\textheight]{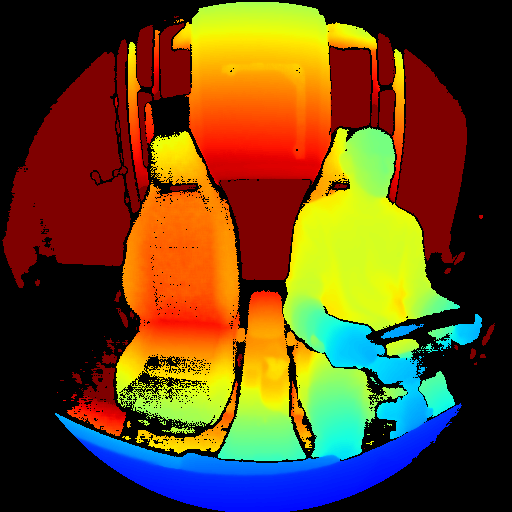}
		\label{fig:h_up}
	\end{subfigure}
	\begin{subfigure}[b]{0.33\textwidth}
		\centering
		\includegraphics[height=0.2\textheight]{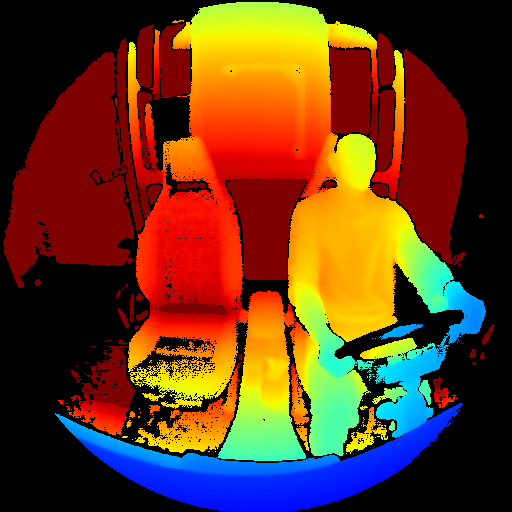}
		\label{fig:p_back}
	\end{subfigure}
	\begin{subfigure}[b]{0.33\textwidth}
		\centering
		\includegraphics[height=0.2\textheight]{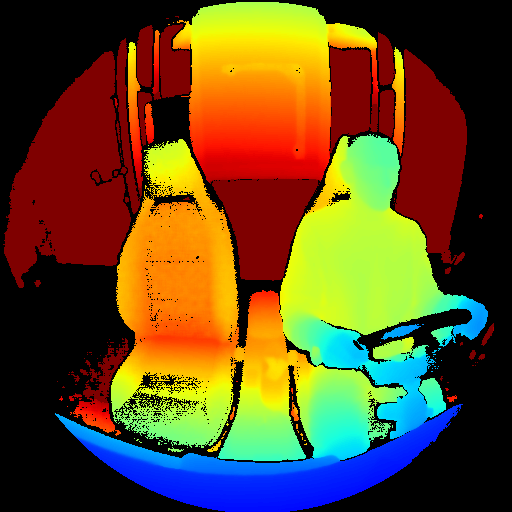}
		\label{fig:p_front}
	\end{subfigure}
	\begin{subfigure}[b]{0.33\textwidth}
		\centering
		\includegraphics[height=0.2\textheight]{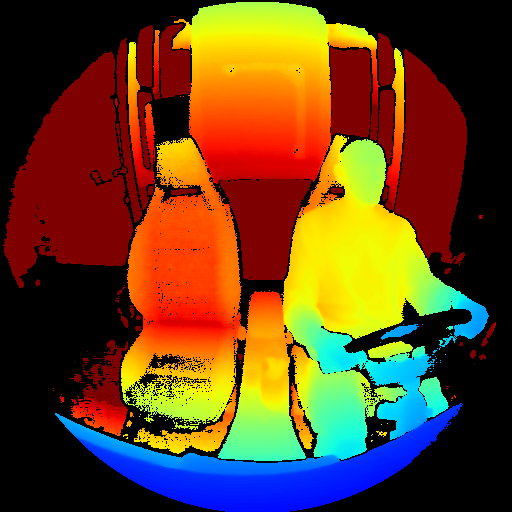}
		\label{fig:t_down}
	\end{subfigure}
	\begin{subfigure}[b]{0.33\textwidth}
		\centering
		\includegraphics[height=0.2\textheight]{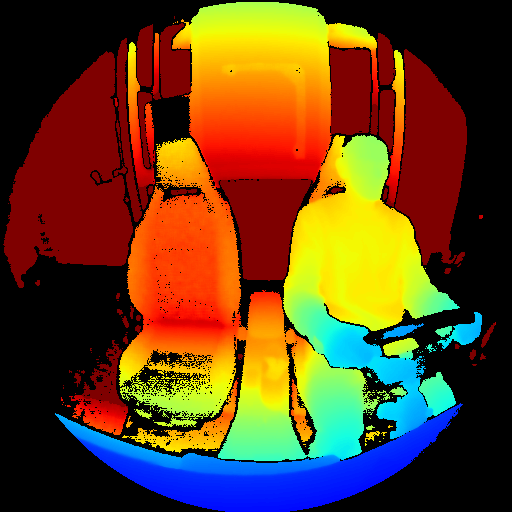}
		\label{fig:t_up}
	\end{subfigure}
	\caption{Depth map comparison of extreme seat positions. In these images, the depth of each pixel is represented in a false-color map ranging from far away (red) to very close (blue). Each row shows the two extreme positions of each seat positioning parameter, 
with the others kept at mean value. From top to bottom: backrest, height, position, tilt.}
	\label{fig:seat_extr}
\end{figure}

As discussed in Section \ref{ss:seatpos}, the position of the mock-up seats can be varied over the full range of four degrees of freedom. In order to demonstrate the influence of the seat positioning on the depth image, we have recorded a scene with a driver and an empty seat with a set of pre-defined seat positions (see Fig. \ref{fig:seat_extr}). For that purpose, each of the four parameters was set to both extreme positions while the others were kept at mean positions. One recognizes that the backrest inclination (top row in Fig. \ref{fig:seat_extr}) has a large influence on the depth image. The measured depth values, as well as the scale and shape of both driver and empty seat, change significantly when under variation of the backrest inclination. Varying the height or position of the seat changes mainly the depth and thus the scale of the objects in the image, while a change of the tilt of the seat base has almost no influence on the depth image. These observations are important for designing an optimal test matrix covering the full range of scene variations with a manageable amount of test cases.

\section{Proposed Usage}
\subsection{Seat occupancy classification}
By monitoring the entire vehicle cabin, the DFKI test platform is supporting the development
of multi-seat occupancy detection, including occupant classification, driver recognition, and object detection functions. 
The robustness of novel deep learning approaches for occupancy classification based on augmented and fully 
synthetic data will be tested against the variability of the scene in driving scenarios.

\begin{figure}
	\centering
	\begin{subfigure}[b]{0.475\textwidth}
		\centering
		\includegraphics[height=0.65\textwidth]{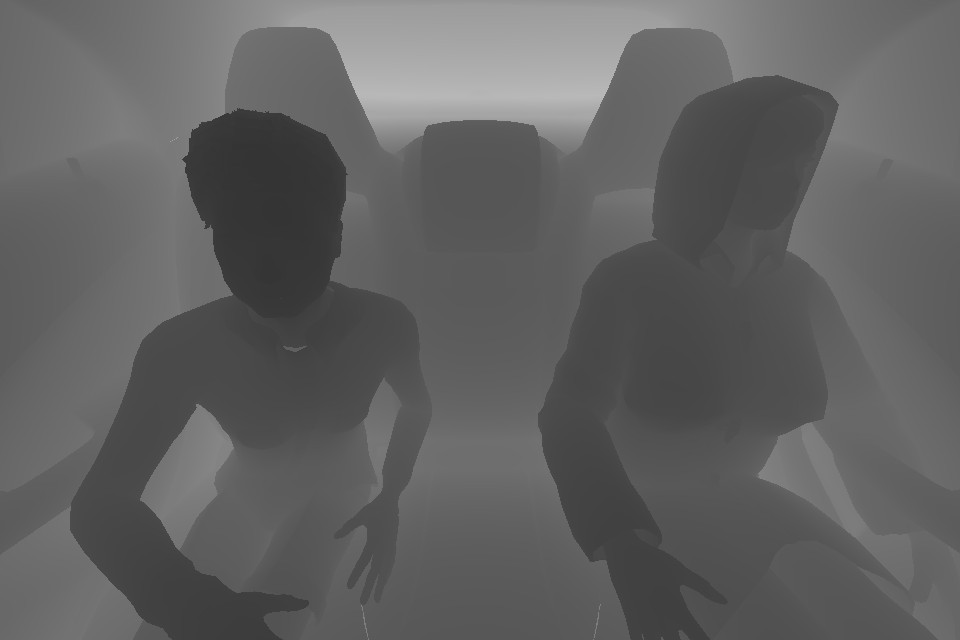}
		\caption{Depth Image from SVIRO \cite{DiasDaCruz2020SVIRO} dataset}
		\label{fig:j_1}
	\end{subfigure}
	\begin{subfigure}[b]{0.475\textwidth}
		\centering
		\includegraphics[height=0.65\textwidth]{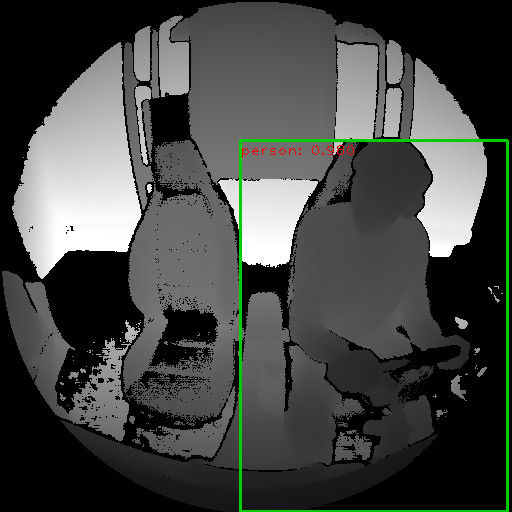}
		\caption{Recorded depth image}
		\label{fig:j_2}		
	\end{subfigure}
	\caption{Depth image from SVIRO dataset (left) with detected person on an image recorded with test platform (right). 
	The depth values are coded as gray values, the detection is indicated by the green bounding box.}
	\label{fig:j}
\end{figure}

We use SVIRO \cite{DiasDaCruz2020SVIRO}, a large-scale synthetic dataset that provides multi-modal image data for scenarios within the passenger compartment of ten different vehicles. Fig. \ref{fig:j} shows an example depth image from the dataset. To show the promise of our test platform, we have trained a simple object detector for the task of detecting people on SVIRO synthetic depth images and evaluated it on real depth images captured in our driving simulator. Our model is based on popular Faster R-CNN \cite{Ren_2017} architecture with all training parameters the same as in the original paper. We used 10k synthetic images for training and let the model train until no further decrease in the loss was observed. To test our model, we recorded a driving scenario with one person in the driving seat in our driving simulator. We achieve a mean AP score of 88.6\% for class 'person' with our simple detector. Fig. \ref{fig:j_2} shows the results of our trained detector on an example image captured with our test platform. 

\subsection{AutoPOSE - Driver head pose ground truth acquisition}

\begin{figure}
	\centering
	\begin{subfigure}[b]{0.475\textwidth}
		\centering
		\includegraphics[height=0.65\textwidth]{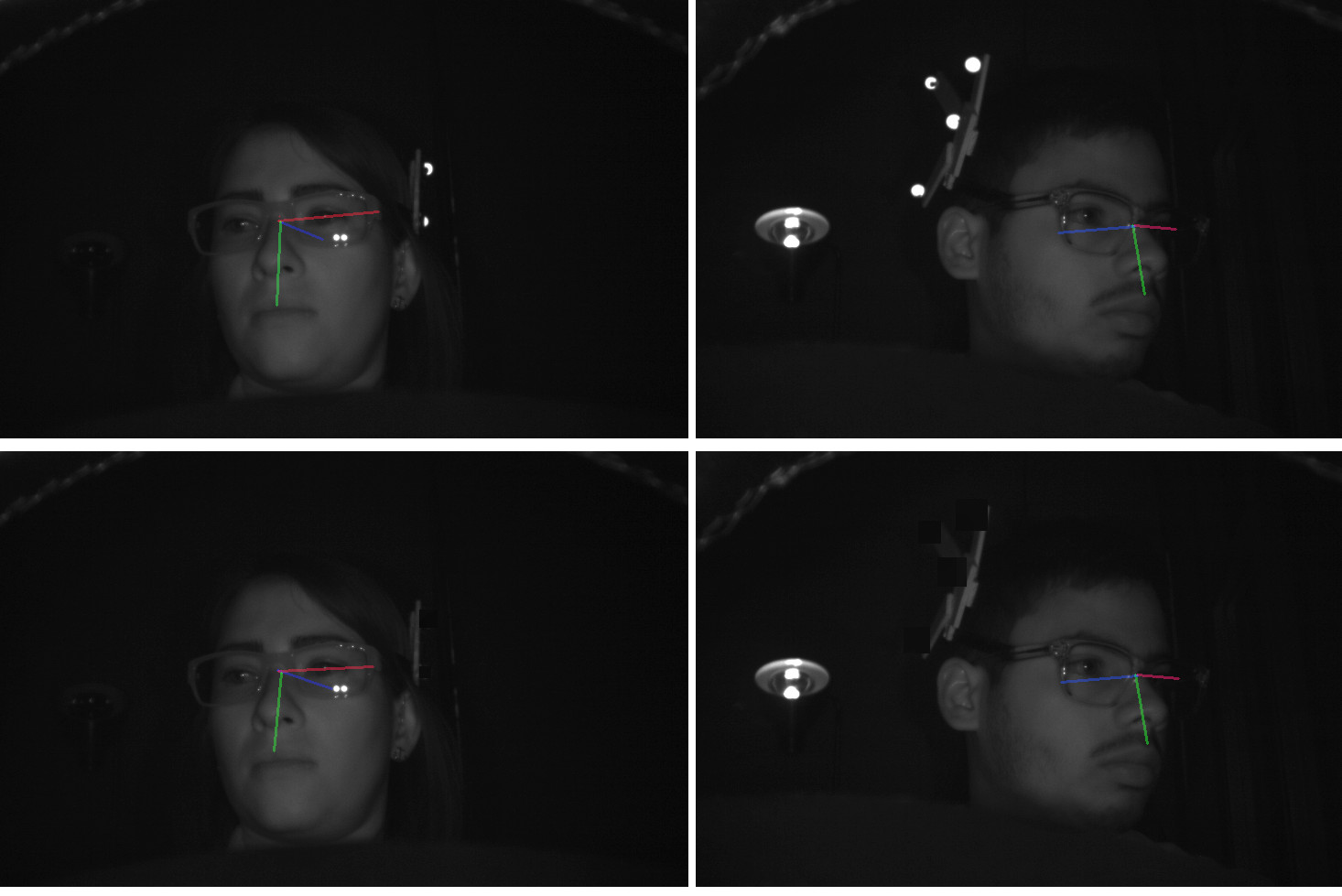}
		\caption{IR Camera - Dashboard}
		\label{fig:j1}
	\end{subfigure}
	\begin{subfigure}[b]{0.475\textwidth}
		\centering
		\includegraphics[height=0.65\textwidth]{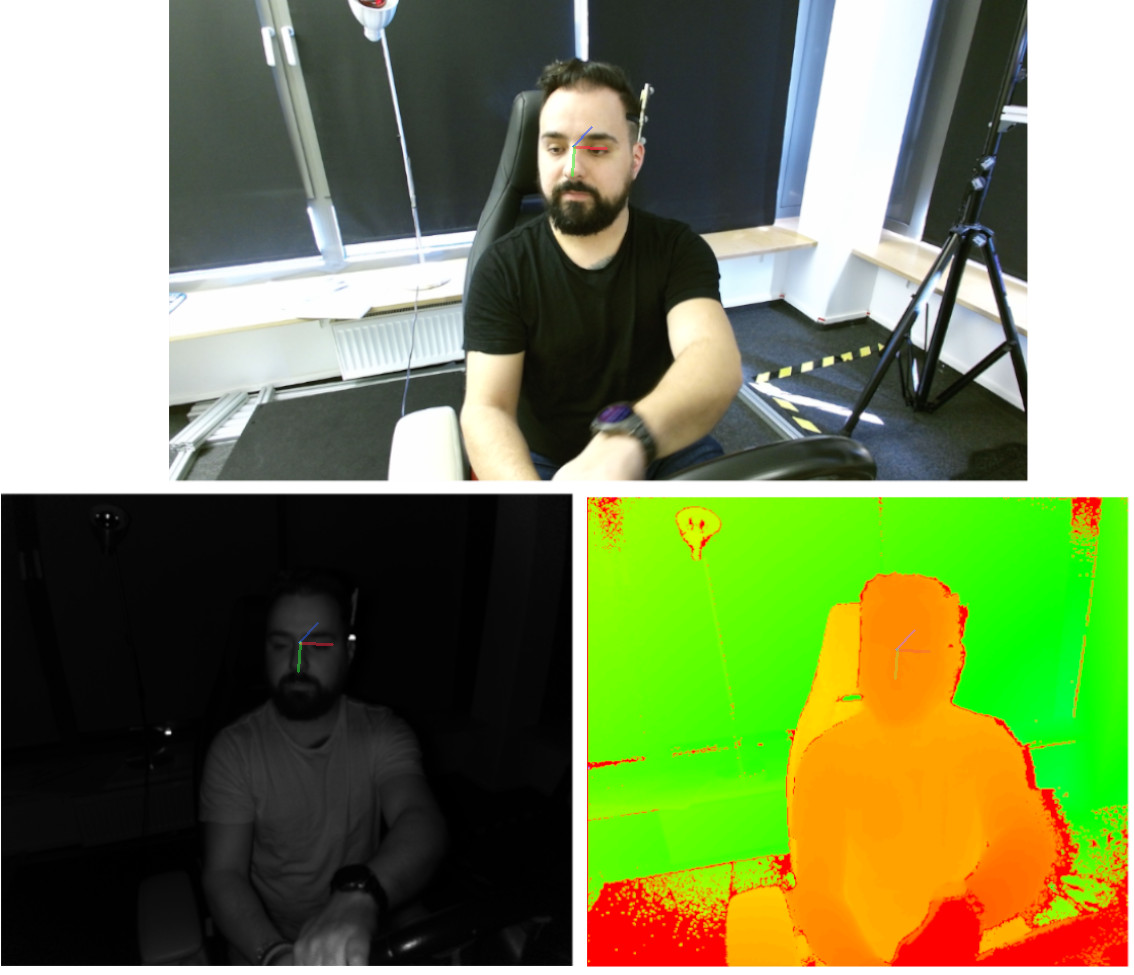}
		\caption{Kinect v2 \cite{kinectv2} - Center mirror}
		\label{fig:j2}		
	\end{subfigure}
	\caption{Sample dataset images from \cite{autoPose}. \\\hspace{\textwidth}
		(a) Row1: RAW images with head target reflective markers visible, Row 2: post-processing - markers covered (accurately localized).
		Second column shows gaze annotation lamp. \\\hspace{\textwidth}
		(b) Kinect v2 \cite{kinectv2} color, IR, and depth (color mapped) images.
		Note: Intensity was improved for visibility and printing purposes.}
	\label{fig:mhd}
\end{figure}

The introduced driving simulator was used in the acquisition of the AutoPOSE dataset \cite{autoPose}. AutoPOSE is a new head pose and eye gaze targets dataset. An IR camera was located at the driver's dashboard, giving almost a frontal view of the driver's face. A Kinect v2 was placed at the location of the center mirror (rear mirror) of the car providing 3 image types, IR, depth, and RGB images. The subjects total number was 21 (10 females and 11 males). The dashboard IR camera subset consists of \textbf{1,018,885} IR images and the Kinect subset consists of \textbf{316,497} synchronized RGB, depth, and IR images. As mentioned in Section \ref{ss:optitrack}, we used the submillimeter accurate OptiTrack motion capturing system for accurate and reliable tracking data acquisition that can be used as ground truth. The driver's head coordinate system and the cameras were calibrated and synchronized with the tracking system. Besides having the driver's head pose, also the eye gaze targets were acquired. The subjects were asked to gaze at reflective markers placed at driving-related locations, for example, side mirrors, center rear mirror, dashboard, road view, and media center. Fig. \ref{fig:mhd} shows samples of the dataset from the IR camera perspective and the Kinect camera perspective. Please refer to the dataset paper for further details.

\subsection{3D body pose tracking}
3D body pose tracking is the basis for a comprehensive activity recognition of the vehicle occupants, and also supports a robust analysis of the driver's state and availability. Since it is possible to equip the subject with as many tracking markers as needed, a broad range of applications can be tested with individual joint positioning.

\begin{figure}[H]
	\begin{center}
		\includegraphics[width=0.4\textwidth]{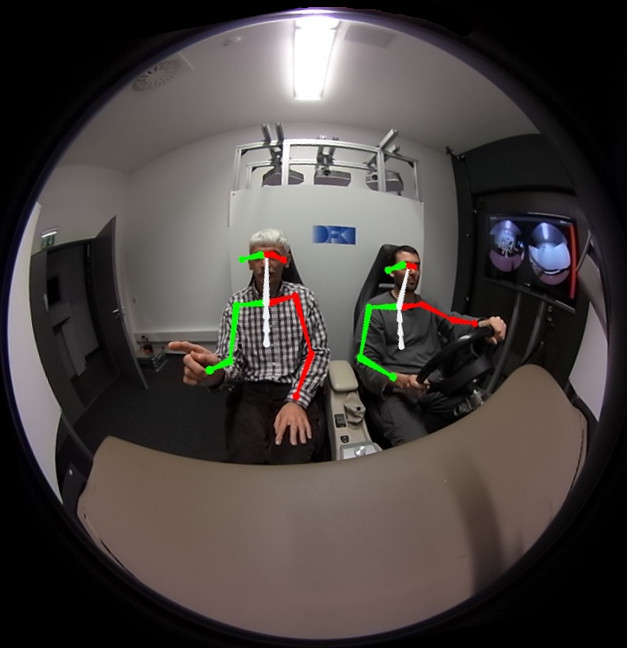}
		\caption{Running 3D pose tracking algorithm in the driving simulator}
		\label{fig:pose_algo}
	\end{center}
\end{figure}

A first deep learning approach towards a full 3D body pose tracking in a car based on a spherical camera was published in \cite{pub9349} and \cite{ElhayekKMMS18} and was already demonstrated in real-time in the driving simulator (Fig. \ref{fig:pose_algo}). Comprehensive recording of driving scenarios with ground truth data will allow to quantitatively evaluate these algorithms.

\subsection{Gesture, activity, and intention recognition}
Another recent research activity has been the development of novel deep-learning-based hand pose and gesture
recognition in a car \cite{pub9138}. It was shown that special recurrent neural networks allow the recognition
of more complex hand gestures than what is currently available as a state-of-the-art sensor system. 
The setup for this study was a time-of-flight camera installed in the rear mirror of a vehicle. 
The test platform allows an extension of this study towards the recognition of arm gestures and activities. 
The test platform offers moreover the possibility to investigate approaches to fuse the visual information 
with other sensor modalities, as, e.g. vehicle driving parameters, for a robust intention recognition of the driver.

\subsection{Benchmark dataset}
In the future, the DFKI in-cabin test platform will be used to record and annotate a large-scale database of in-cabin 
scenes under realistically simulated driving scenarios. The database will support the benchmark testing of a range of monitoring
functions as those described above, in the first line seat occupancy classification, occupant size, and 3D-pose estimation,
as well as driver activity and intention recognition. Apart from the ground truth data recording for the pose estimation, 
the data needs to be annotated with various, application-specific attributes, including also a 2D- and 3D- labeling of the object
location, size, and orientation. Therefore, semi-automated annotation tools are under development. 

\section{Conclusion}
In this paper, we presented a multi-purpose in-cabin test platform. It provides a near-realistic driving experience and is equipped with a 2D/3D camera setup and a ground truth sensor system. Thus it can be used to accurately evaluate driver monitoring algorithms and also to create in-cabin datasets. The construction was explained in detail, various use cases have been described and first results were shown. We lay great emphasis on high adaptability for different scenarios and we hope to use this platform for years to come in various projects.

\section*{Acknowledgement}
\label{s:acknowledgement}

This project has received funding within the Electronic Components and Systems
for European Leadership (ECSEL) Joint Undertaking in collaboration with the
European Union's H2020 Framework Program and National Authorities, under
grant agreement n° 826600 (VIZTA).

\bibliographystyle{splncs03}
\bibliography{references_av_cvt2020}

\end{document}